\documentclass[conference]{IEEEtran}
\usepackage[utf8]{inputenc}
\usepackage{amsmath}
\usepackage{amsfonts}
\usepackage{amssymb}
\usepackage{graphicx}
\usepackage{algpseudocode}
\usepackage{flushend}
\usepackage{xspace}
\usepackage[colorlinks = true,
            citecolor = blue,
            bookmarks=false
            ]{hyperref}
\usepackage{footnote}
\usepackage{subfigure}
\usepackage{algorithm}
\usepackage{bbm}

\newcommand{\ie}{\emph{i.e.}, }
\newcommand{\eg}{\emph{e.g.}, }

\newcommand{\ada}{\emph{AdaComp}\xspace}
\newcommand{\comp}{\emph{Comp-ASGD}\xspace}

\begin{document}

\title{Distributed deep learning on edge-devices:\\ feasibility via adaptive compression}

\author{\IEEEauthorblockN{ Corentin Hardy}
\IEEEauthorblockA{\textit{Technicolor}, \textit{Inria}\\
Rennes, France}
\and
\IEEEauthorblockN{Erwan Le Merrer}
\IEEEauthorblockA{\textit{Technicolor}\\
Rennes, France}
\and
\IEEEauthorblockN{Bruno Sericola}
\IEEEauthorblockA{\textit{Inria} \\
Rennes, France}
}

\maketitle

\begin{abstract}
A large portion of data mining and analytic services use modern machine learning techniques, such as deep learning. 
The state-of-the-art results by deep learning come at the price of an intensive use of computing resources. The leading frameworks (\eg
TensorFlow) are executed on GPUs or on high-end servers in
datacenters.  On the other end, there is a proliferation of personal
devices with possibly free CPU cycles; this can enable services to run in users' homes, embedding machine learning operations.
In this paper, we ask the
following question: \textit{Is distributed deep learning computation
  on WAN connected devices feasible, in spite of the traffic caused
  by learning tasks}?  We show that such a setup rises some important
challenges, most notably the ingress traffic that the servers
hosting the up-to-date model have to sustain.

In order to reduce this stress, we propose \ada, a novel algorithm for
compressing worker updates to the model on the server. Applicable to stochastic gradient descent
based approaches, it combines efficient gradient selection and
learning rate modulation.  We then experiment and measure the impact
of compression, device heterogeneity and reliability on the accuracy of learned
models, with an emulator platform that embeds 
TensorFlow into Linux containers.
We report a reduction of the total amount of data sent by workers to the
server by two order of magnitude (\eg $191$-fold reduction for a
convolutional network on the MNIST dataset), when compared to a standard
 asynchronous stochastic gradient descent,
while preserving model accuracy.
\end{abstract}

\section{Introduction}
\label{sec:intro}

Machine learning methods, and in particular deep learning, are nowadays key
components for building efficient applications and services. Deep
learning recently permitted significant improvements over state-of-the-art techniques for building classification models for
instance~\cite{Imagenet_DNNClassif_2012}. Its use spans over a large spectrum of applications,
from face recognition in~\cite{DeepFaceReco_BMVC2015}, to natural language processing~\cite{Word2Vec_NIPS2013} and to video recommendation in YouTube~\cite{Youtube_recom_application_2016}.

Learning a model using a deep neural network (we denote DNN hereafter)
requires a large amount of data, as the precision of that model
directly depends on the quantity of examples it gets as
input and the number of times it iterates over them. 
Typically, the last image recognition DNNs, such as~\cite{DeepImageReco_2016} or~\cite{Imagenet_DNNClassif_2012}, leverage very large datasets (like Imagenet~\cite{Imagenet_dataset}) during the learning phase; this leads to the processing of over $10$TB of data. 
The direct consequence is the compute-intensive nature of running such approaches. 
The place of choice for running those methods is thus well provisioned datacenters,
for the more intensive applications, or on dedicated and GPU-powered
machines in other cases. In this context, recently introduced
frameworks for learning models using DNNs are
benchmarked in cloud environments (\eg TensorFlow~\cite{osdi_tf} with
GPU-enabled servers and 16Gbps network ports).

\begin{figure}[t!]
\centering
\includegraphics[scale=0.32]{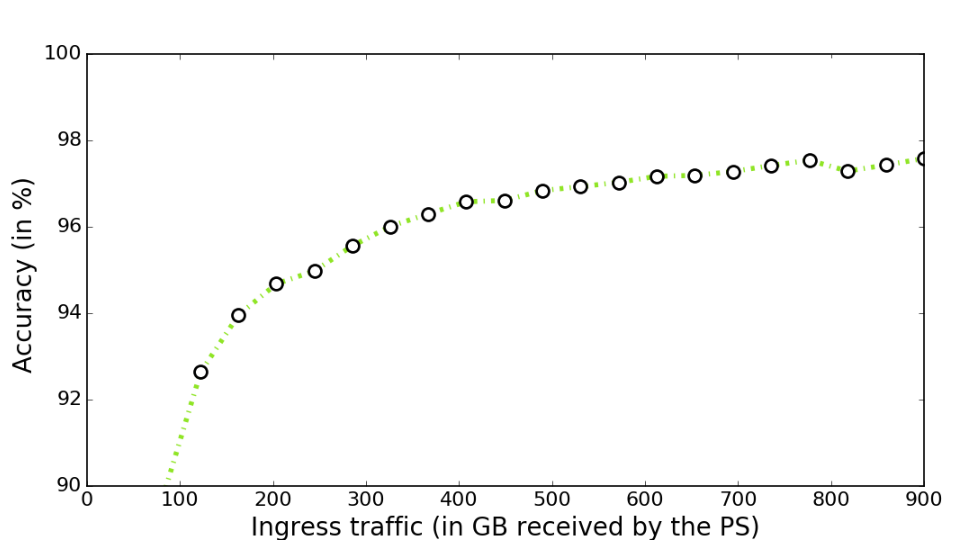}
\caption{Accuracy (in \%) vs aggregated traffic (in GB) tradeoff: inbound aggregated ingress traffic received at the Parameter Server (PS). For an asynchronous stochastic gradient descent with 200 workers, a convolutional neural network model training, on the MNIST dataset.}
\label{aggregateddata}
\end{figure}

Meanwhile, the number of processing devices at the edge of the
Internet keeps increasing in a steep manner. In this paper, we explore
the possibility of leveraging edge-devices for those intensive
tasks. This new paradigm rises significant feasibility questions.  The
dominant \textit{parameter server} computing model, introduced by
Google in 2012~\cite{DistBelief_DownpourSGD_2012}, uses a set of
\textit{workers} for parallel processing, while a few central servers (denoted the \textit{parameter server} (PS) hereafter for simplicity)  are managing shared states modified by those workers. Workers frequently
fetch the up-to-date model from the PS, make computation over
the data they host, and then return gradient updates to the
PS. Since DNN models are large (from thousands to
billions parameters~\cite{Benchmark_Framework_2016}), placing those
worker tasks over edge-devices imply significant updates transfer over
the Internet. The PS being in a central location (typically at a
cloud provider), the question of inbound traffic is
also crucial for pricing our proposal. Model learning
 is facing other problems such as device crashes and worker asynchrony.
We note that those concerns differ from \textit{federated optimization} techniques~\cite{FederatedLearning_Comm_2016}, that aim at removing worker asynchrony to reach minimal communication, but at the cost of removing processing parallelism that motivated the PS approach in the first place.

To illustrate the feasibility question, we implement the largest
distribution scenario considered in the TensorFlow
paper~\cite{osdi_tf}, where 200 machines are collaborating to learn a
model. We measure the aggregated traffic at the PS generated by the learning of a classifier on the MNIST dataset, and
plot it on Figure~\ref{aggregateddata}.
We observe a considerable amount of ingress traffic received by the PS, of the order of Terabyte for an accurate model. This amount of traffic is due to workers sending their updates to the PS, and is not even reported in research studies, as well provisioned and dedicated
data center networks are assumed.
Clearly, in a setup leveraging edge-devices, this amount of ingress traffic has to be
drastically reduced to remove weight on both the Internet and on
the PS. Meanwhile, the data to be processed at
edge-devices themselves is not the limiting factor ($3.9$MB each in this experiment, as the dataset is split among workers).
Our solution is to introduce a novel compression technique for sending
updates from workers to the PS, using 
gradient selection~\cite{PrivacyPreservingDDL_Shokri_2015}. We thus study the model accuracy with regards to update compression, as well as with regards to device reliability.

The main contributions of this paper are the following:
\textit{1)} Exposing the parameter server model implications, in an edge-device setup. 
\textit{2)} Introducing \ada, a novel compression technique for reducing the ingress traffic at the PS. We detail \ada formally, and also open-source its code for public use.\footnote{The code is available on \href{https://github.com/Hardy-c/AdaComp}{https://github.com/Hardy-c/AdaComp}} 
\textit{3)} Experimenting \ada within TensorFlow, and comparing it to competitors with regards to model accuracy.

First, in Section~\ref{sec:psmodel}, we briefly present the basics of deep
learning, and of distributed learning in datacenters.
Section~\ref{sec:ddlmodel} introduces the considered execution setup 
on edge-devices, and important performance metrics. Section~\ref{sec:adacomp} presents \ada, before evaluating it  and its competitors
in Section~\ref{sec:evaluation}. We present related work in Section~\ref{sec:related} and concluding remarks in Section~\ref{sec:conclusion}.

\section{Distributed Deep Learning} 
\label{sec:psmodel}

\subsection{Basics on DNN training on a single core}
\label{subsec:sgd}
DNNs are machine learning models used for supervised or unsupervised tasks. They are composed of a large succession of
layers. Each layer $l$ is the result of a non-linear transformation of
the previous layer (or input data) given a weight matrix $V_l$ and a
bias vector $\beta_l$. 
In this paper, we focus on supervised learning, where a DNN is used to approximate a target function $f$ (\eg a classification function for images).

For instance, given a training dataset $\mathcal{D}$ of images $x_h \in X$ and associated labels $y_h \in Y$, a DNN has to adapt its parametric function $\hat{f}_{\Theta} : X \rightarrow Y$, with $\Theta$ a vector containing the set of all parameters, \ie all entries of matrices $V_l$ and vectors $\beta_l$. Training is performed by minimizing a loss function $L$ which represents how good is the DNN at approximating $f$, given by : 

\begin{equation}
L(\Theta) = \sum_{(x_h,y_h)\in\mathcal{D}} e(y_h,\hat{f}_{\Theta}(x_h)),
\end{equation}

\noindent where $e(y_h,\hat{f}_{\Theta}(x_h))$ is the error of the DNN output $\hat{f}_{\Theta}(x_h)$ for $x_h$ with parameters $\Theta$. The minimization of $L(\Theta)$ is performed by an iterative update process on the parameters called gradient descent (GD). At each step, the algorithm updates the parameters as follows (vector notations) :

\begin{equation}
\Theta(i+1) = \Theta(i) - \alpha \Delta \Theta(i),
\end{equation}

\noindent with the learning rate $\alpha \in (0,1]$ and $\Delta \Theta(i)$ approximating the gradient of the error function. $\Delta \Theta(i)$ is computed by a \textit{back-propagation} step
\cite{lecun-98b} on $\mathcal{D}$ at time $i$. In the following sections, we use the more efficient variant of GD, called Stochastic Gradient Descent (SGD) \cite{bottou-98x}, which processes a mini-batch of the training dataset per iteration.

\begin{figure}
\centering
\includegraphics[scale=0.4]{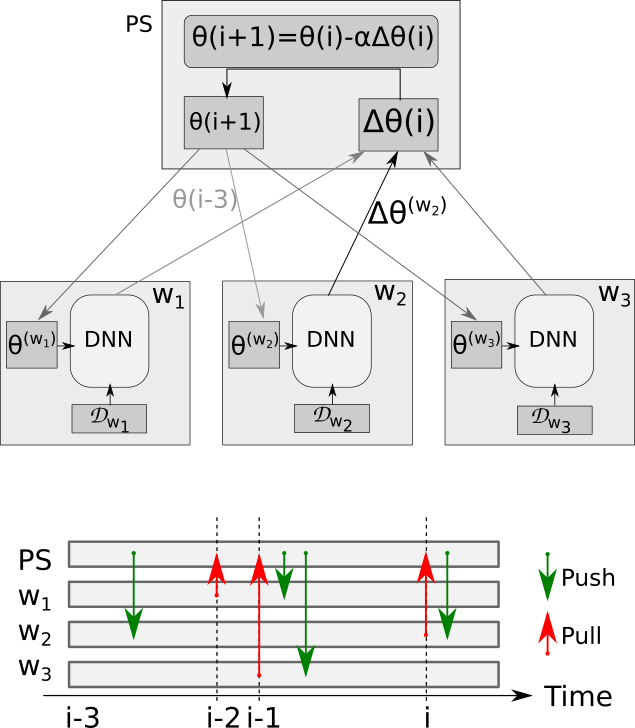} 
\caption{The asynchronous \textit{parameter server} computation model, with 3 workers and an update chronology.
  }\label{PS_datacenter}
\end{figure}

\subsection{Distributed Stochastic Gradient Descent in the Parameter Server Model}

To speed-up DNN training, Dean et al. proposed in
\cite{DistBelief_DownpourSGD_2012} the \textit{parameter server} model, as a way to distribute computation
onto up to hundreds of machines. The principal idea is to parallelize data computation: each compute node (or \textit{worker})
processes a subset of the training data asynchronously. For convenience, we report the core notations in
Table~\ref{tab:notations}.

\begin{savenotes}
\begin{table}[t!]
  \centering 
  \caption{List of notations.}\label{tab:notations}
\begin{tabular}{|p{16.5mm}|p{60mm}|}
\hline
PS & The parameter server, hosting the up-to-date $\Theta$\\
$n$ & The number of workers (excluding the PS)\\
$w_1,\dots,w_n$ & The identifiers of the $n$ workers\\
$b$ & Size of mini-batches\\ 
$\alpha$ & The model learning rate\\
$c$ & The compression ratio \\& ~(\% of selected parameters [per $V_l$ or $\beta_l$])\\
$\mathcal{D}_w$ & Training data local to a worker $w$\\ 
$I$ & The number of performed iterations\\  
$\Theta(i)$ & Vector containing all parameters at iteration $i$,\ie all entries of matrices $V_l$ and vectors $\beta_l$ at $i$\\
$\Theta_k(i)$ & The $k$-th entry of $\Theta(i)$\\
$\Delta \Theta(i)$ & Vector containing all updates to $\Theta$\\ 
$\sigma(i)$ & The global staleness of update $i$\\
$\sigma_{k}(i)$ & Local update staleness, associated to parameter $\Theta_{k}$\\
$\alpha_{k}(i)$ & Local learning rate associated to $\Theta_{k}$ given $\sigma_{k}(i)$\\
$\Theta^{(w)}$, $\Delta \Theta^{(w)}$ and $\Delta \tilde{\Theta}^{(w)}$ & The local parameter vector state at worker $w$, the update computed from $\Theta^{(w)}$ by worker $w$ and the resulting compressed update\\
\hline
\end{tabular}

\end{table}
\end{savenotes}

The parameter server model is depicted on
Figure~\ref{PS_datacenter}. In a nutshell, the PS starts by
initializing the vector of parameters to learn, $\Theta$. Training data $\mathcal{D}$ is split across  workers $w_1,w_2,w_3$. Each worker runs asynchronously, and whenever it is ready, takes the current version of the parameter vector from the PS, performs a SGD step and sends the vector update (that reflects the
learning on its data) to the PS.  
In Figure~\ref{PS_datacenter}, worker $w_2$ gets the state of $\Theta$ at iteration $i-3$, performs a SGD step and sends an update to the PS. 
Meanwhile, workers $w_1$ and $w_3$, which finish their step before $w_2$, send their updates.
At each update reception, the PS
updates $\Theta$, and increments the timestamp of $\Theta$ (\eg
$\Theta(i-3)$ becomes $\Theta(i-2)$ after the reception of $w_1$'s
update). An important parameter is the mini-batch size, denoted by
$b$, which corresponds to the subset size of the local data that each
worker is computing upon, before sending one update to the PS (\eg if
$b=10$, an update is sent by $w_1$ after its DNN has processed $10$ images of $\mathcal{D}_{w_1}$). 

Note that in the PS model, the fact that
$n$ workers are training their local vector in parallel and then send
the updates introduces concurrency, also known as
\textit{staleness}~\cite{Staleness-awareASGD_2015}.  The
staleness of the local vector $\Theta^{(w)}$ for a worker $w$ is the number of times $\Theta$
has evolved in between the fetch by $w$, and the
time where it itself pushes an update $\Delta \Theta^{(w)}$ (\eg on Figure~\ref{PS_datacenter}, $w_2$'s update staleness is equal to $2$).


\section{Distributed Deep Learning  on Edge-Devices}
\label{sec:ddlmodel}


The execution setup we consider replaces the datacenter worker nodes
by edge-device nodes (\eg personal computers of
volunteers with SETI@home, or home gateways~\cite{Valancius2009}) and the
datacenter network by the Internet. The PS remains at a
central location.
In this context, we assume that the training data reside with the workers; this serves for instance as a basis for privacy-preserving scenarios~\cite{PrivacyPreservingDDL_Shokri_2015}, where users have their photos at home,
and want to contribute to the computing of a global photo
classification model, but without sending their personal data to a
cloud service.

The formal execution model remains identical to the PS model: we suppose that workers have enough memory to host a copy of $\Theta$, data is shuffled on the $n$ workers, and finally they
agree to collaborate to the machine learning task (tolerance to malicious behaviours is out of the scope of this paper). The stringent aspect of
our setup is the lower connectivity capacity of workers, and their
expected smaller reliability~\cite{Valancius2009}. Regarding connectivity, we have in mind
a standard ASDL/cable setup where the bottleneck at the device is
the uplink (with \eg respectively 100Mb/10Mb for down/up bandwidth);
we thus optimize on the worker upload capacity by compressing the updates
it has to send \ie the push operation, rather than the opposite \ie the pull operation.  

\subsection{Staleness mitigation}
\label{subsec:stalness}
Asynchronous vector fetches and updates by workers involve perturbations on the SGD due to the staleness of local $\Theta^{(w)}$. The staleness, proportional to the number of workers (please
refer to \cite{Rudra_2015}, \cite{AsyncSGD_NIPS2015} or \cite{Staleness-awareASGD_2015} for in depth phenomenon explanation), is an important issue in asynchronous SGD in general.
A high staleness has to be avoided, as workers that are relatively slow will contribute,
through their update, to a $\Theta$ that has evolved (possibly
significantly) since they last fetched their copy of $\Theta$. This
factor is known for slowing down the computation
convergence~\cite{Staleness-awareASGD_2015}.
In order to cope with it, works~\cite{FasterASGD_2016} and \cite{Staleness-awareASGD_2015} adapt the learning rate $\alpha$ as a function of the current staleness to reduce the impact of stale updates, which will also reduce the number of updates needed to train~$\Theta$.

\subsection{Reducing ingress traffic at the PS}
\label{sub:reducing}

In the edge-device setting, where devices collaborate to the
computation of a global $\Theta$, we argue that the critical metric is
the bandwidth required by the PS to cope with incoming worker
updates. This allows for frontend server/network dimensioning, and is
also a crucial metric, as cloud providers often bill on upload
capacities to cloud servers (see \eg Amazon Kinesis, which charges
depending on the number of 1MB/s message queues to align as
frontends). Since our setup implies best effort computation at
devices, not to saturate uplink, the workers are sending updates to
the PS as background tasks. We thus measure the total ingress traffic at
the PS collection point, in order to have an aggregated view of the
upload traffic from workers, incurred by the deep learning tasks.

\paragraph{Compression} Shokri \textit{et al.}
\cite{PrivacyPreservingDDL_Shokri_2015} proposed a compression
mechanism for reducing the size of the updates sent from each worker
to the PS, named \textit{Selective Stochastic Gradient Descent}. At
the end of an iteration over its local data, a worker sends only
a subset of computed gradients, rather than all of them. The
selection is made either randomly, or by keeping only the largest
gradients. The compression ratio is represented by fixed value $c \in (0,1]$.
They experimentally show that model accuracy is not impacted
much by this compression method for a SGD in a single core (with \eg a MLP model accuracy decreasing from
98,10\% to 97,07\% accuracy, on the MNIST dataset and a selection of
 1\% of parameters per update).

In the light of this Section, the total amount of ingress traffic received at the PS is of the order of $I \times M \times c$, with $M$ the size of $\Theta$ (\eg in MB).
As the crucial focus of machine learning is the accuracy of the learned model, we have shown (Figure~\ref{aggregateddata}) that unfortunately it is not linear with the amount of data received at the PS. That is why we 
measure the accuracy/ingress traffic tradeoff experimentally, in the remaining of this paper.
As we shall see in the evaluation Section, approach~\cite{PrivacyPreservingDDL_Shokri_2015} in an edge-device setup manages to reduce the size of updates, but at the cost of accuracy. This is a clear impediment, because deep learning is leveraged for its state-of-art accuracy results. In order to cope with both ingress traffic and accuracy maintenance, we now introduce the \ada algorithm.

\section{Compressed updates with \ada}
\label{sec:adacomp}

\ada is a solution to conveniently combine the two concepts of
compression and staleness mitigation, for further compressing worker
results. This permits drastic ingress traffic reduction, for a
comparable accuracy with best related work approaches.

\paragraph{Rationale}
To do so,  we propose the following approach. We first
observe that the content of updates pushed by workers are often
\textit{sparse} (\ie most of the parameters have not changed, $\Delta \Theta_k \approx 0$ for most of $\Theta_k$): a selection method based on largest values of update is a good solution to compress it with little loss of information. Second, we observe that staleness mitigation is handled solely at the
granularity of a whole update, in related approaches. 
From those remarks, we
 propose to compress updates pushed by worker and use staleness mitigation. The novelty in \ada is to compute staleness not on an update, but
\textit{per parameter}. Our intuition is that sparsity and staleness
mitigation on individual parameters will allow for increased efficiency in
asynchronous worker operation, by removing a significant part of
update conflicts. An independent staleness score computed for each parameters is a reasonable assumption: F. Niu \textit{et al.} \cite{Hogwild_NIPS2011} consider parameters as independent during the SGD.

\paragraph{Selection method} We
propose a novel method for selecting gradient at each worker: only a fraction
$c$ of the largest gradients per matrix $V_l$ and vector $\beta_l$ are
kept. This selection permits to balance learning across DNN
layers and better reflects computed gradients as compared to a random selection (or solely the largest ones across the whole model) as in~\cite{PrivacyPreservingDDL_Shokri_2015}.

\paragraph{Algorithm details}
\ada operates in the following way, also described with pseudo-code in
Algorithm~\ref{algo:worker} and \ref{algo:ps}. The PS keeps a trace of
all received updates at given timestamps (as in~\cite{Rudra_2015,Staleness-awareASGD_2015}). When a worker pulls $\Theta$, it receives the associated timestamp $j$. 
It computes a SGD step and uses the selection method to compress the update.
Meanwhile, intermediate updates (pushed by other workers) increase
the timestamp of the PS to $i\geq j$. Instead of computing the same staleness for each $\Delta \Theta_k(i)$ of the update $\Delta \Theta(i)$, we define an adaptive staleness for each parameter $\Theta_k$ as follows~:

\begin{equation}\label{eq:sigma}
\sigma_{k}(i) = \sum_{u=j}^{i-1} \mathbbm{1}_{\{\Delta \Theta_k(u) \neq 0 \}},
\end{equation}
where $\mathbbm{1}_A$ is the indicator function of condition $A$ equal to $1$ if 
condition $A$ is true and $0$ otherwise.
The staleness $\sigma_{k}$ is then computed \textbf{individually} for each parameter
by counting the number of updates applied on it since the last pull by the worker.
We use the update equation inspired by \cite{Staleness-awareASGD_2015}:

\begin{equation} \label{eq:update}
\Theta_k(i+1) = \Theta_k(i) - \alpha_{k}(i) \Delta \Theta_k(i),
\end{equation}

\noindent where

\begin{equation}\label{eq:alpha}
\alpha_{k}(i) =  
	\begin{cases}
	\alpha/\sigma_{k}(i) &\mbox{ if } \sigma_{k}(i) \neq 0  \\
	\alpha &\mbox{otherwise.}
	\end{cases}
\end{equation}

The $\alpha_{k}$ is the learning rate computed for the $k$-th
parameter of $\Theta$ given staleness $\sigma_{k}$. 
The parameter $\Theta_k$ that was updated since the
worker pull, will see $\alpha_{k}$ reduced as a function of the
staleness, as shown in equation (\ref{eq:alpha}). This method has the
effect of reducing the concurrency between the possibly numerous asynchronous updates
taking place along the learning task, while taking advantage of gradient selection to reduce update size\footnote{Please note that the best low level data representation for update encoding is out of the scope of this paper: it will causes at best a reduction of a small factor w.r.t. python serialization (\ie the TensorFlow language), while we target in this paper two orders of magnitude decrease on the network footprint by targeting the algorithmic scope of distributed deep learning.}.

\begin{algorithm}[t!]
\begin{algorithmic}[1]
\Procedure{Worker}{$PS,\mathcal{D}_w,I,b,c$} 
    \State $i \gets 0$
	\While{$i<I$}
		\State $\Theta^{(w)},i \gets $ Pull\_$\Theta(PS)$ \Comment{get current $\Theta(i)$ from PS}
		\State $\Delta \Theta^{(w)} \gets$ SGD\_STEP$(D_w,b)$ 
        \State $\Delta \tilde{\Theta}^{(w)} \gets$ SELECT\_GRAD$(\Delta \Theta^{(w)},c)$
		\State Push$(\Delta \tilde{\Theta}^{(w)},i)$ \Comment{update push to PS, with fetched i}
		
	\EndWhile \label{Worker}
\EndProcedure
\Procedure{SGD\_STEP}{$\mathcal{D}_w,b$} 
    \State \textit{Select a mini-batch $B$ by sampling $b$ examples from $\mathcal{D}_w$ and compute $\Delta \Theta^{(w)}$ with back-propagation method.}
    \State \textbf{return} $\Delta \Theta^{(w)}$
\EndProcedure
\Procedure{SELECT\_GRAD}{$\Delta \Theta^{(w)},c$} 
    \State \textit{Select $\Delta \tilde{\Theta}^{(w)} \subset \Delta \Theta^{(w)}$ by keeping the $(100\times c)\%$ largest parameters, in absolute value,  of 
    each matrix $V_l$ and vector $\beta_l$.}
    \State \textbf{return} $\Delta \tilde{\Theta}^{(w)}$
\EndProcedure
\end{algorithmic}
\caption{\ada at workers}\label{algo:worker}
\end{algorithm}

\begin{algorithm}
\begin{algorithmic}[1]
\Procedure{PS}{$\alpha,I$} 
	\State \textit{Initialize  $\Theta(0)$ with random values.}
	\For{ $i \gets 0,I$}	
		\State Get$(\Delta\tilde{\Theta}^{(w)},j)$ \Comment{wait for a push}
		\State $\Delta\Theta(i) \gets \Delta\tilde{\Theta}^{(w)}$ 
		\ForAll{$\Delta \Theta_k(i)\in \Delta \Theta(i)$}
			\State $\sigma_{k} \gets \sum_{u=j}^{i-1} \mathbbm{1}_{\{\Delta \Theta_k(u) \neq 0 \}}$ 
			\State \textbf{if} $\sigma_{k}=0$ \textbf{then} $\alpha_{k} \gets \alpha$ \textbf{else} $\alpha_{k} \gets \alpha / \sigma_{k}$
			\State $ \Theta_k(i+1) \gets \Theta_k(i) - \alpha_{k} \Delta \Theta_k(i)$
		\EndFor
	\EndFor\label{euclidendwhile}
\EndProcedure
\end{algorithmic}
\caption{\ada at the parameter server}\label{algo:ps}
\end{algorithm}

\paragraph{Complexity } \ada implies increased complexity for the PS to compute the adaptive staleness $\sigma_k$ for each $\Theta_k$. In terms of memory, the PS has to maintain 
the last $d$ updates, where $d$ is the worst delay (\ie the delay of the last worker which did not sent an update). Only indexes of non-zero parameters are maintained, leading to a memory complexity equals to $O(d \times c \times | \Theta | )$. In the worth-case, the computation of $\sigma_k$ requires $d$ dichotomous searches in lists of sizes $c \times | \Theta |$ (assuming that indexes of each update are stored in a sorted list). The computational complexity of $\sigma_k$ is $O(d \times \ln(| \Theta | \times c))$ for a parameter, and this computation occurs for $c \times | \Theta | $ parameters at each iteration. We thus obtain an overall complexity of  $O( d \times c \times | \Theta | \times \ln(| \Theta | \times c) ) $ per iteration for the PS. Note that $d$ has the same magnitude as $n$ (see \cite{Staleness-awareASGD_2015}), so we can rewrite the complexity as $O( n \times c \times | \Theta | \times \ln(| \Theta | \times c) ) $. 

Note that in the classical PS model the computation complexity is $O( | \Theta |)$; we thus conclude in a computation complexity increase by $O(n \times c \times \ln(| \Theta | \times c) )$ in \ada, highlighting the proposed computation vs communication tradeoff. Finally for workers, the selection of the largest parameters does not increase the complexity.

\section{Experimental Evaluation}
\label{sec:evaluation}


\subsection{Experimental platform}

For assessing the value of \ada in a controlled and monitored
environment, we choose to emulate the overall system on a single
powerful server. The server allows us to remove the hardware and
configuration constraints. We choose to represent edge-devices through
Linux containers (LXC) on a Debian high end server (Intel(R) Xeon(R) CPU
E5-2667 v3 @ 3.20GHz CPUs, a total of 32 cores, and 1/2TB of
RAM). Each of them runs a TensorFlow session to train locally the DNN (we note that TensorFlow is also running in
  containers, while executed in a datacenter
  environment). The traffic between LXCs can then be
managed by the host machine with \textit{virtual Ethernet} (or veth)
connections.
We recall the open-sourcing of the algorithm code (please refer to footnote 1). 

An experiment on this platform is as follows.  A set of $n$ LXC
containers is deployed, to represent the workers. Each worker in a
container has access to a proportion $1/n$ of the training dataset.  One
LXC container is deployed to run the PS code.  All workers are
connected to PS by a veth virtual network. Finally, one last LXC
container is deployed to evaluate the accuracy evolution of $\Theta$.

During the execution, the platform adds a random waiting time, for
each worker, between the effective computation time and the push step
to the PS. This time ensures that the order of worker updates is
random, mimicking the possible variety in hardware or resource
available to the workers (we furthermore conduct an experiment on this device heterogeneity in the experiment Section). 
We report a runtime of about  $1,5$ day for each run (\ie for reaching
$I=250,000$ iterations).

In such a setup, the interesting metric to observe is the accuracy
reached according the number of iterations performed, which is the
same, whether computation takes place in an emulated setup or in a
real deployment.  The benefit of our platform is a tight monitoring
of the resulting TCP traffic, during the learning task.

\subsection{Experimental setup and competitors}
\label{sec:expsetup}
We experiment our setup with the MNIST dataset~\cite{lecun1998mnist}, also used by our competitor~\cite{PrivacyPreservingDDL_Shokri_2015}. The goal is to
build an image classifier able to recognize handwritten digits (\ie 10
classes). The dataset is composed of a training dataset with $60,000$
images, \ie $6,000$ per class, and a test dataset with $10,000$
images. Each image is represented by $28\times28$ pixel with a 8-bit grey-level matrix.

We experiment in this paper on a Convolutional Neural Network (CNN), consisting of two convolutional layers and two full-connected layers ($211,690$ parameters),
and is taken from the Keras library \cite{chollet2015keras} for the
MNIST dataset. 
Our tech report~\cite{tech-rep} additionally contains experiments over a Multi-Layers Perceptron model.
The experiments are launching $n=200$ workers, and one PS,
corresponding to the scale of operation reported by TensorFlow~\cite{osdi_tf}.

We compare the performance of \ada to the ones of 1) the basic
Async-SGD method (which we denote ASGD)~\cite{AsyncSGD_NIPS2015} as a baseline.
2) the \comp algorithm, which is similar to ASGD, but implements a gradient selection as described in 
\cite{PrivacyPreservingDDL_Shokri_2015}. For both competitors, the learning rate is divided by staleness and is global to each update.


\subsection{Accuracy results}

\begin{figure*}[t!]
  \begin{minipage}{0.48\linewidth}
    \centering
\includegraphics[scale=0.38]{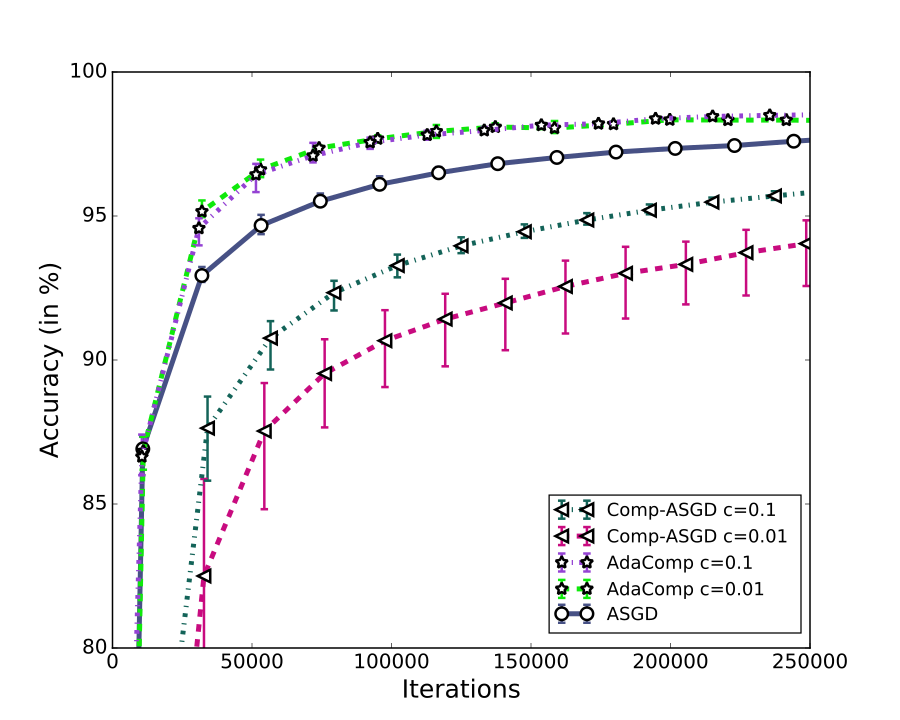} 
  \end{minipage}\hspace{0.15cm}
  \hspace{0.05cm}
  \begin{minipage}{0.48\linewidth}
    \centering
\includegraphics[scale=0.38]{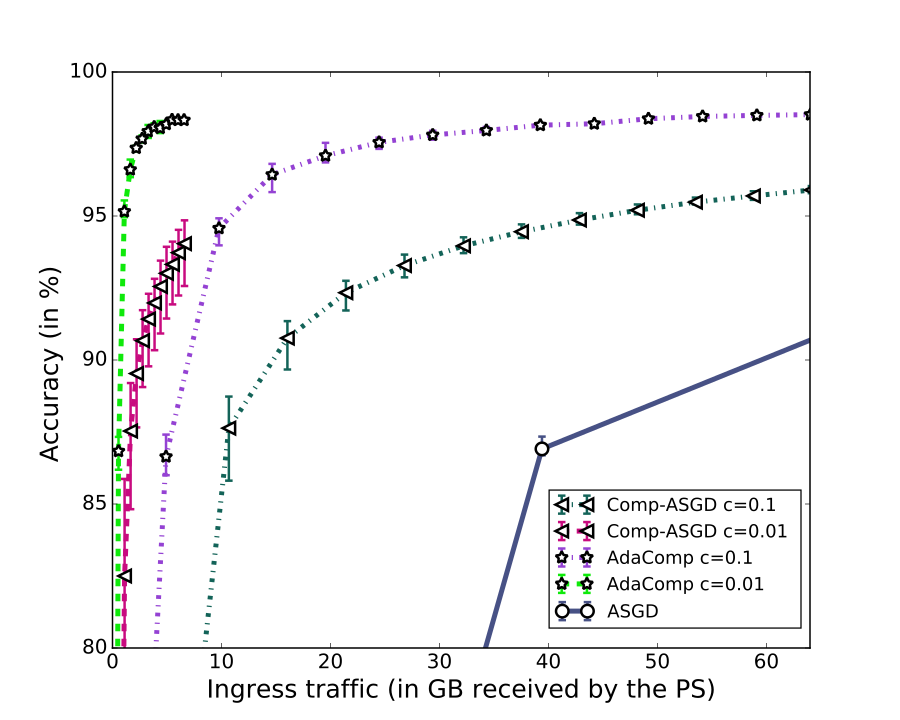} 
  \end{minipage}%
  \caption{CNN model learning in the edge-device setup: accuracy
    results (Left), and resulting ingress traffic (Right).} 
   \label{fig:CNN}
\end{figure*}

All experiments are executed until the PS has received a total of
$I=250,000$ updates from the workers; this corresponds to an upper
bound for convergence of the best performing approaches we present in
this Section. Experiments have been run 3 times; plots show
average, minimal and maximal values of the moving average accuracies for each run. Final accuracy is the mean of the maximal accuracy reached by each run.

Figure \ref{fig:CNN} (left) plots the accuracy of the model depending on the number of iterations.
Figure \ref{fig:CNN} (right) reports the accuracy as function of the ingress traffic, measured at the PS.
ASGD, \comp and \ada are experimented
within the same setup with a batch size : $b=10$. For both
\comp and \ada, we test the compression ratios $c = 0.1$ and $c=0.01$,
respectively representing $10\%$ and $1\%$ of the
local parameters sent as an update by a worker to the PS.

Figure~\ref{fig:CNN} (left) first shows that 
compression affects \comp results, while it does not
prevent \ada to perform better than both ASGD and \comp.  Final
accuracy results are for ASGD $97.85\%$, for \comp $96.07\%$ and $95.11\%$ ($c =
0.1$ and $0.01$ respectively), and finally for \ada $98.7\%$ and $98.59\%$.
We report in \cite{tech-rep} that lower values for $c$ (equals to $0.001$ or $0.0001$), result in degraded accuracy.

Regarding the resulting ingress traffic by considered algorithms,
presented on Figure~\ref{fig:CNN} (right), we observe striking
differences. As expected, not relying on compression causes ASGD to
produce a large amount of traffic in destination to the PS (up to
$460$GB after $250,000$ iterations), while the accuracy is still not at its possible best. The effect of compression for \comp and \ada, with low values of $c$,
is clearly noticeable. This allows to drastically reduce the
amount of traffic the PS has to deal with, and then in turn reduces the
operational cost of the deep learning task.

As a mean for precise comparison, we fix an accuracy level of $97\%$, and
report the resulting aggregated ingress traffic for each algorithm. \ada generates $1.33$GB of ingress traffic with $c=0.01$, representing $191$ times less traffic than ASGD. \comp does not manage to reach this accuracy level.

We conclude that \ada outperforms ASGD and \comp on both accuracy and ingress traffic (consistently with the second model experimented in \cite{tech-rep}). While compressing the size of updates naturally leads to a reduced ingress traffic at he PS, \ada beating ASGD is less
 intuitive: \ada counters the effect of staleness
due to the asynchronous updates, by using fine grained updates. The probability of update conflicts of a parameter $\Theta_k$ is indeed less important than at the level of
the whole set of parameters $\Theta$ in an update (as performed in \comp). This allows for a
higher learning rate $\alpha_k$ for parameters not concerned by staleness.
This makes it possible to consider DNN computation on edge-devices (\eg for a home gateway connected 24/7, this experiment corresponds to an average of $~5$KB/s upload over a week).


\subsection{\ada accuracy facing worker failures}

A key element for accuracy of distributed deep learning is the reliability of computation facing device failures.
We then consider fail-stop type of failures for workers (\eg they crash without warning
messages or partial updates to the PS). 
In addition, we consider that the local data of a crashed node are lost for the system.
We tune the probability $p=0.004$ that each worker crashes after each push operation. We then simply freeze the randomly selected containers, to emulate crashes. In expectation, half of the initial population of $200$ workers will have crashed by the end of the experiment (\ie $200$ workers are present at $I=0$,
and only $100$ survived at $I=250, 000$). Results for \ada with $c=0.01$ 
are presented on Table~\ref{table:crashes}. We operate on
the MLP presented in our tech report~\cite{tech-rep}.

The first observation, with \ada $c=0.01$ with crashes,
is that crashes have very little effect on accuracy ($97.17\%$, \ie $0.27\%$ less). This is due to the fact that the MNIST dataset is "too rich" for the learning task: this translates by nodes crashing with not mandatory data for the accuracy of $\Theta$ in the end.

\begin{table}[t!]
   \centering
   \caption{Maximal accuracy of \ada, with $c=0.01$ and half the workers crashing during the experiment.}
   \begin{tabular}{l | c | c | c}
     \hline
     \multicolumn{3}{l|}{Parameters} & Reached accuracy\\
     \hline
     Training set & Crashes & n\\
     \hline
     \hline
      $60, 000$ images & no & 200 & $97.44\%$\\
      $60, 000$ images & yes & 200 & $97.17\%$\\
      $12, 000$ images & no & 200 & $95.47\%$\\
      $12, 000$ images & yes & 200 & $95.24\%$\\
      $6, 000$ images & no & 100 & $94.35\%$\\
     \hline
    \end{tabular}
    
\label{table:crashes}
\end{table}

We run the same experiment with only $20\%$ of the original MNIST dataset as training
set (\ie $12,000$ images).
We report a loss of accuracy of $0.23\%$ between \ada with crashes and \ada with no crashes (with $c=0.01$ for both), close to the previous loss of accuracy with the whole training dataset.
To further explain this phenomenon, we run an experiment where only $100$ workers participate to the learning task. We test \ada for $c=0.01$, $n=100$, $10$\% of the training set and no crash  (\ie equivalent to a scenario where half of the 200 nodes would have disappeared with thus $10\%$ of data prior to the start of the learning task). Accuracy is $94.35\%$, versus $95.24\%$ for the previous experiment with crashes. This underlines that the $100$ crashed nodes have in fact participated to some extent to the model learned (otherwise \ada with crashes and $20\%$ of training set would also have terminated around $94.35\%$ accuracy as well).

This experiment underlines that crash failures of edge-devices will not affect the accuracy of the model if the dataset over which they learn is rich enough, and that the impact of failures remains very limited otherwise (assuming little contribution of devices).

\subsection{\ada accuracy facing heterogeneous workers}

For each edge-device, the joint effect of its network constraints, hardware and concurrently executed tasks could lead to significantly varying latencies in the computation of a batch. 
Distributed learning tasks then have to cope with that worker heterogeneity of completion times.

To assess the performance of \ada facing this heterogeneity, this experiment builds on three different classes of workers: a fast class, a medium and a slow one. Fast workers send 10 times more updates to the PS than medium workers, and 100 time more than slow workers. The proportion of workers in each class is respectively $30\%$, $40\%$ and $30\%$ (workers do not switch class during runtime). We run \ada with parameters $c=0.001$, $n=200$. 
We experiment on a \textit{large training set}, composed by the MNIST training dataset (\ie with $60,000$ images) and a reduced training set composed by $20\%$ of the MNIST dataset (\ie $12,000$ images).

\begin{figure}[t!]
\hspace{-0.7cm}
\begin{minipage}{0.48\linewidth}

\includegraphics[scale=0.42]{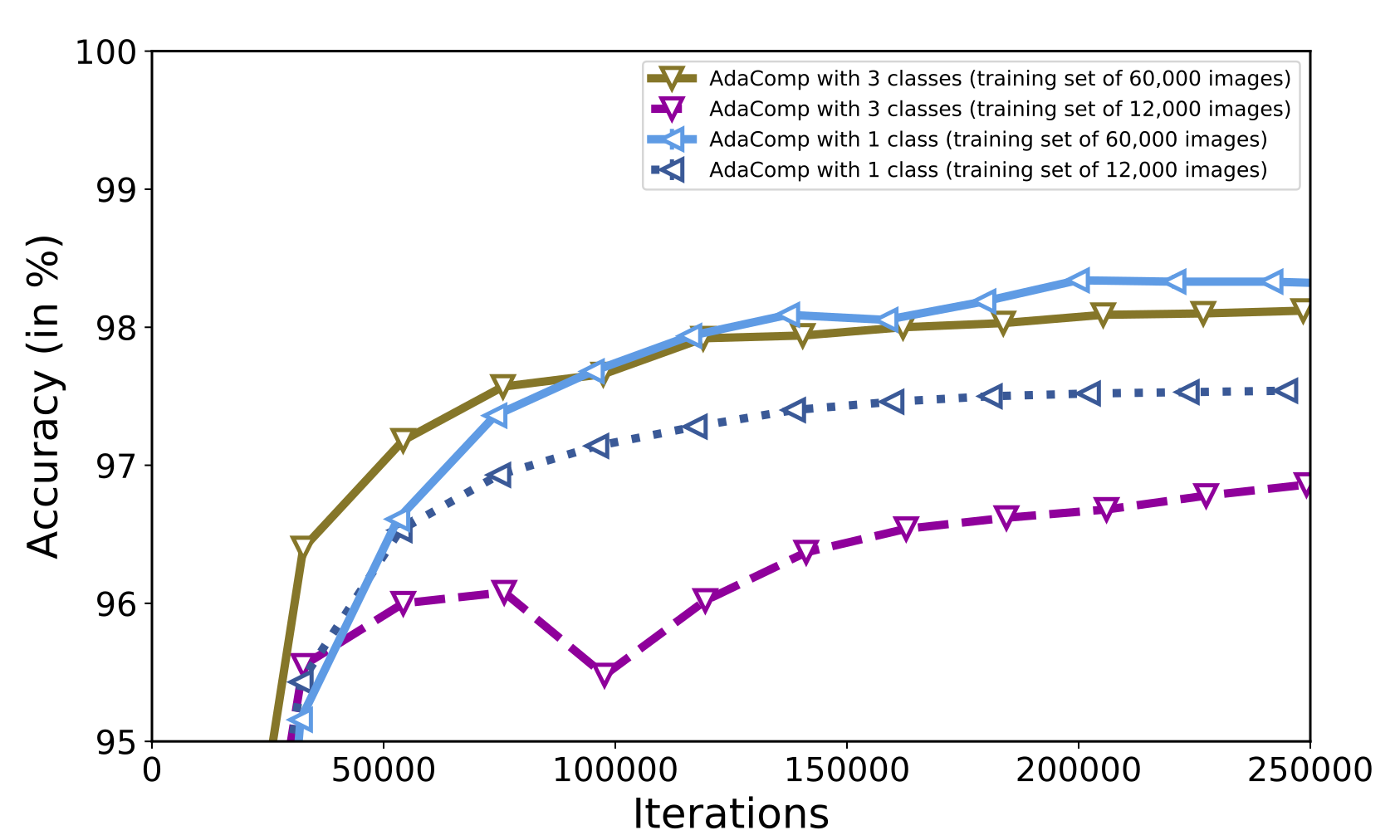}
\end{minipage}
\caption{CNN model learning with three classes of workers (heterogeneous devices).}
\label{fig:hw}
\end{figure}

Figure \ref{fig:hw} plots the accuracy of the model described in \ref{sec:expsetup} (then with solely one class of workers) and with the three classes of workers. 
The first observation is that \ada with only one class of workers causes a higher final accuracy, as compared to three classes (respectively $98.59\%$ versus $98.36\%$ for the \textit{large training set} and $97.66\%$ versus $97.00\%$ for the \textit{reduced training set}). Note that the gap of accuracy is larger when the training set is reduced. Figure \ref{fig:hw} shows that \ada with three classes of workers converges faster at the beginning than \ada with one class (experiment with \textit{large training set}). This underlines that, with three classes, the fastest workers contribute more to the learning in a shorther amount of iterations, quickly increasing the accuracy at the beginning (with a low staleness score per update). However, when the DNN is adapted to the fastest workers updates, it does not then learn enough from slower workers to reach the same accuracy than the experiment with one class. This phenomenon is particularly salient on the reduced training set curves.

This experiment shows that distributed learning with heterogeneous workers, as expected in real deployments, is feasible as convergence occurs. 
It tends to reduce the contribution of the slowest workers as compared to that of the faster ones. 
Such as in the previous experiment with crash failures, the impact of heterogeneity on the final accuracy depends on the processed dataset (impact is reduced when using a richer training set).


\section{Related Work}
\label{sec:related}

There are numerous alternatives to classical SGD to speed-up deep learning such as Momentum SGD \cite{rumelhart1988learning}, Adagrad \cite{duchi2011adaptive}, or Adam \cite{adam}. Those methods use adaptive gradient descent for each parameter, but are meant to run on a single machine (\ie are not suitable for distributed computing), and then are not competing with \ada for parallel speed-up. Future works may adapt Adam with \ada to speed-up the learning in a distributed setup.

The parameter server model is popular for distributing SGD
with \eg DistBelief~\cite{DistBelief_DownpourSGD_2012}, Adam
\cite{Project-Adam_OSDI2014}, and TensorFlow~\cite{osdi_tf} as the principal ones. Increasing the number of workers in such an asynchronous
environment causes staleness problems, that has been addressed either \textit{1)} algorithmically, or by means of \textit{2)} synchronization.

\textit{1)} In \cite{Staleness-awareASGD_2015}, W. Zhang et \textit{al.} propose to adapt the learning rate according to the staleness for each new update. The PS divides the learning rate by the computed staleness of the worker. This method limits the impact of highly stale updates. More recently, Odena~\cite{FasterASGD_2016} proposed to maintain the averaged variance of each parameters to precisely weight new updates. Combined with the previous method, this allows to adapt the learning rate for each parameter according the changes provided by previous workers. The learning rate will be higher for parameters which witnessed few changes than those which witnessed big changes during the last updates. This method is close to \ada, but does not take into account sparse updates.

\textit{2)} A simple solution to avoid staleness problems is to synchronize worker updates. However, waiting for all workers at each iteration is time consuming. W. Zhang et al.~\cite{Staleness-awareASGD_2015} propose the $n$-softSync protocol where the PS waits a fraction $1/n$ of all workers at each iteration before updating parameters. A more accurate gradient is computed by averaging the computed gradient of the fraction of workers. Another recent work of Chen \textit{et al.}~\cite{SyncSGD_2016} shows that a synchronous SGD could be more efficient if the PS does not wait the last $k$ workers at each iteration. 
In our setup, workers are user-devices without any guarantee on the upper bound of their response time. This calls for efficient asynchronous methods like \ada.

Finally, at the other extreme of the parallelism versus communication tradeoff, so-called \textit{federated optimization} has been introduced~\cite{FederatedLearning_Comm_2016,FederatedOptim_2016,FederatedLearning_premices}.
A model is learned given a very large number of edge-devices each only processing over a few data, and each being equipped with a poor connection. A subset of active workers (which changes at each iteration) runs many iterations over workers data before their model is shared. Global updates are then performed synchronously.
Federated learning is thus communication-efficient, but does not take advantage of data parallelism for speeding-up computation,  which is the goal of the parameter server model we target, with the proposed  \ada algorithm.


\section{Conclusion}
\label{sec:conclusion}
We discussed in this paper the most salient implications of
running distributed deep learning training tasks on edge-devices.
Because of upload constraints of devices and of their lower reliability in
such a setup, asynchronous SGD is a natural solution to perform
learning tasks; yet we highlighted that the amount of traffic that has
to transit over the Internet are considerable, if algorithms are not adapted.
We thus proposed \ada, a new algorithm to
compress updates by adapting to their individual parameter staleness.
We show that this translates into a 
$191$-fold
reduction of the ingress traffic at the parameter server, as compared to the asynchronous SGD algorithm (for 
CNN models on the MNIST dataset), and for an as well
better accuracy.
This large reduction of the ingress traffic, and the reliability facing crashes makes it possible
to consider the actual deployment of learning tasks on edge-devices, for powering edge-services and applications.

\bibliographystyle{IEEEtran}
\bibliography{MDDL}

\end{document}